\def\year{2020}\relax
\documentclass[letterpaper]{article} % DO NOT CHANGE THIS
\usepackage{aaai20}  % DO NOT CHANGE THIS
\usepackage{times}  % DO NOT CHANGE THIS
\usepackage{helvet} % DO NOT CHANGE THIS
\usepackage{courier}  % DO NOT CHANGE THIS
\usepackage[hyphens]{url}  % DO NOT CHANGE THIS
\usepackage{graphicx} % DO NOT CHANGE THIS
\usepackage{graphicx}  % DO NOT CHANGE THIS
\usepackage{amsmath,amsfonts,bm}
\usepackage{booktabs}
\usepackage{algorithm}
\usepackage{algorithmic}
\usepackage{multirow}
\usepackage{breqn}
\usepackage{amsfonts}
\usepackage{amssymb}
\usepackage{bbm}

\def\Eqref#1{Eq.~\eqref{#1}}
\def\Tabref#1{Table~\ref{#1}}

\DeclareMathOperator*{\relu}{ReLU}

\def\vb{{\bm{b}}}

\def\vs{{\bm{s}}}

\def\vw{{\bm{w}}}
\def\vx{{\bm{x}}}

\def\mC{{\bm{C}}}

\def\mS{{\bm{S}}}

\def\mW{{\bm{W}}}
\def\mX{{\bm{X}}}

\def\gD{{\mathcal{D}}}

\def\gG{{\mathcal{G}}}

\def\gT{{\mathcal{T}}}

\def\sE{{\mathbb{E}}}
\title{Attribute Propagation Network for Graph Zero-shot Learning }
\author{Lu Liu,\textsuperscript{\rm 1} Tianyi Zhou,\textsuperscript{\rm 2} Guodong Long,\textsuperscript{\rm 1}\thanks{Corresponding author} Jing Jiang,\textsuperscript{\rm 1} Chengqi Zhang\textsuperscript{\rm 1} \\
\textsuperscript{\rm 1} Center for AI, School of Computer Science, University of Technology Sydney  \\
\textsuperscript{\rm 2} Paul G. Allen Center for Computer Science \& Engineering, University of Washington\\
lu.liu-10@student.uts.edu.au, tianyizh@uw.edu, \{guodong.long, jing.jiang, chengqi.zhang\}@uts.edu.au 
}
 
\begin{document}

\maketitle

\begin{abstract}

The goal of zero-shot learning (ZSL) is to train a model to classify samples of classes that were not seen during training. To address this challenging task, most ZSL methods relate unseen test classes to seen(training) classes via a pre-defined set of attributes that can describe all classes in the same semantic space, so the knowledge learned on the training classes can be adapted to unseen classes. 
In this paper, we aim to optimize the attribute space for ZSL by training a propagation mechanism to refine the semantic attributes of each class based on its neighbors and related classes on a graph of classes.
We show that the propagated attributes can produce classifiers for zero-shot classes with significantly improved performance in different ZSL settings.
The graph of classes is usually free or very cheap to acquire such as WordNet or ImageNet classes.
When the graph is not provided, given pre-defined semantic embeddings of the classes, we can learn a mechanism to generate the graph in an end-to-end manner along with the propagation mechanism. 
However, this graph-aided technique has not been well-explored in the literature. 
In this paper, we introduce the ``attribute propagation network (APNet)'', which is composed of 1) a graph propagation model generating attribute vector for each class and 2) a parameterized nearest neighbor (NN) classifier categorizing an image to the class with the nearest attribute vector to the image's embedding.
For better generalization over unseen classes, different from previous methods, we adopt a meta-learning strategy to train the propagation mechanism and the similarity metric for the NN classifier on multiple sub-graphs, each associated with a classification task over a subset of training classes. In experiments with two zero-shot learning settings and five benchmark datasets, APNet achieves either  compelling performance or new state-of-the-art results.

\end{abstract}

\section{Introduction}
\begin{figure}[t!]
% \vspace{-0.8em}
\begin{center}
\includegraphics[width=0.9\columnwidth]{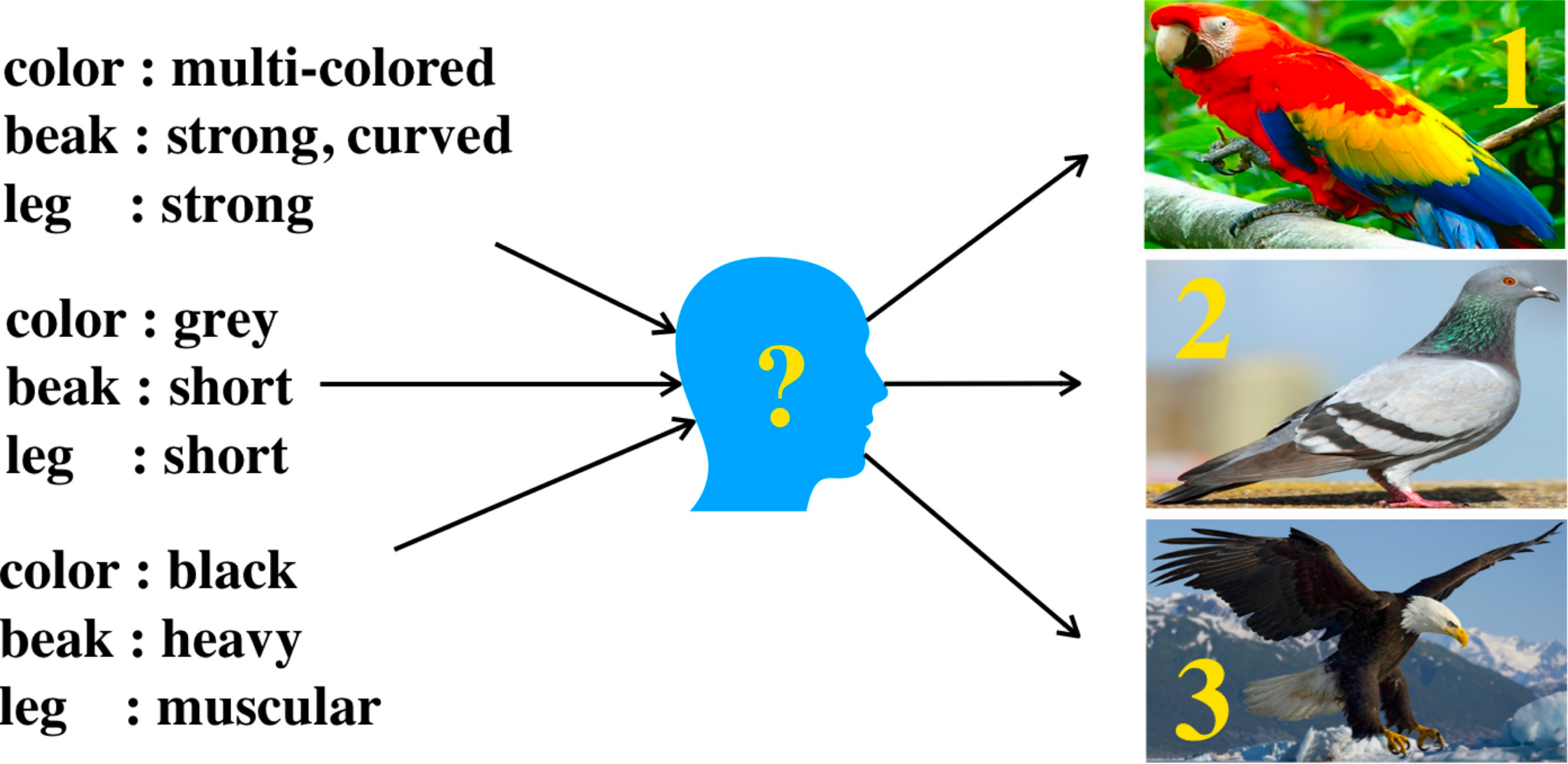}
\end{center}
% \vspace{-1.0em}
\caption{A simple scenario for zero-shot learning. 
It is easy for humans to transform the attribute information into mental pictures and choose the correct image. Can machines achieve this same level of intelligence?}
\label{fig:zero-shot-simple}
% \vspace{-2.0em}
\end{figure}

Imagination, the ability of synthesizing novel objects and reasoning new patterns from existing ones, plays an important role in human's exploration and learning within the unknown world when supervised information is insufficient. 
% peoples and ideas in the mind without any immediate input of the senses. 
% Given a list of attributes describing different classes, humans can accurately classify an image by sight alone, even if they have never seen samples from that class before. 
Given a list of attributes describing different classes, humans can accurately classify an image, even if they have never seen samples from that class before. 
For example, as shown in Figure~\ref{fig:zero-shot-simple}, it is not difficult for someone to match each image with a given set of attributes. 
However, underneath this seemingly simple task, we are imagining---imagining how colors, shapes, and concepts in images relate to words describing them, e.g., the class names. This raises the question: Can machines also ``imagine''? Can a machine be trained to match text-based attribute representations to visual feature representations? 
% However, underneath this seemingly simple task, we are imagining---imagining how colors, shapes, and concepts in images relate to words describing them, e.g., the class names. This raises the question: Can machines also “imagine”? Can a machine use its imagination to match text-based attribute representations to visual feature representations? 
% Humans can accurately do visual recognition only based on related descriptions, even if they have never seen samples from the category of the descriptions. For example, as shown in Figure.~\ref{fig:zero-shot-simple}, given only some attributes of a bird species, it is not difficult for humans to choose the image of the correct class described by the attributes. Can machines also be able to ``imagine'' the visual feature representations given the attribute representations? 
% Zero-shot learning (ZSL) is a challenging task to test exactly this by establishing a scenario and asking a model to “imagine” unseen classes~\cite{awa1}. 
\begin{figure}[ht!]
\begin{center}
\includegraphics[width=\columnwidth]{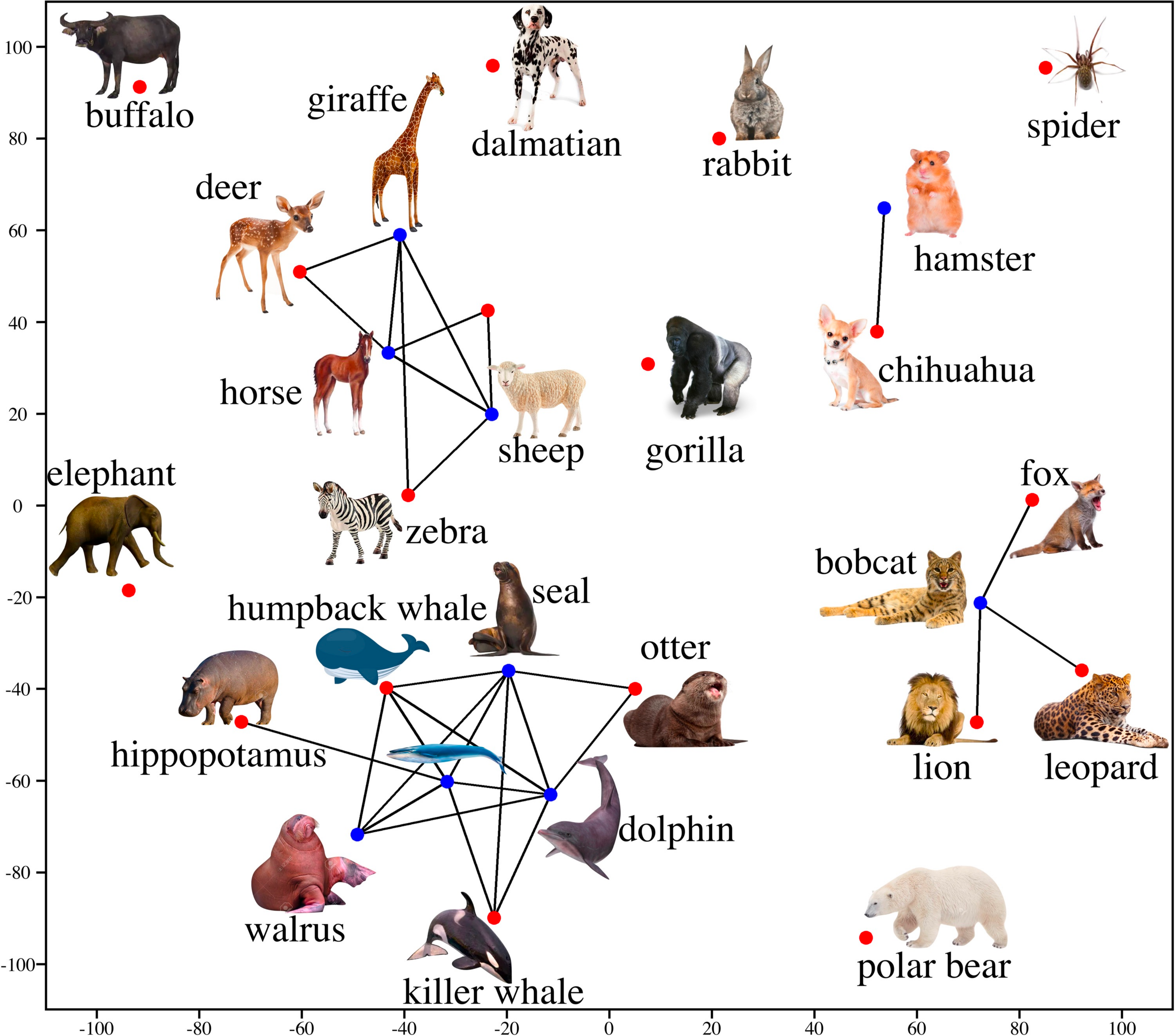}
\end{center}
% \vspace{-1.2em}
\caption{Visualization of the refined attribute vector per class produced by APNet by using t-SNE~\cite{maaten2008visualizing} and the graph of classes generated based on given semantic embedding per class. The red nodes and blue nodes are the propagated attribute vectors for training classes and test classes, respectively. In the scatter plot, the attribute vectors of strongly-related classes are close to each other and these classes are connected by edges on the generated category graph (better viewed in color).}
\label{fig:TSNE-Results}
% \vspace{-1.0em}
\end{figure}

Zero-shot learning (ZSL) is a challenging task aiming to learn a model classifying  images of any unseen classes given only the semantic attributes of the classes. It can test exactly the capability of generalizing knowledge learned on training classes to unseen classes, by establishing a scenario and asking a model to ``imagine'' the visual features of unseen classes~\cite{lampert2013attribute} based on their semantic attributes.
%The most popular scenario is defined as a classification problem over samples from unseen classes during training, given only the semantic information of the test class names. 
Generalized zero-shot learning (GZSL) assumes a more practical scenario. It is still based solely on semantic attributes of test classes with zero training sample, but the test set is composed of data from both seen and unseen classes, and the predictions are not restricted to the scope of unseen classes. The new challenge comes from the imbalance between seen and unseen classes and the possibility that the model mistakenly categorizes unseen class images as seen classes or vice versa. 
It requires the model to be easily adapted to unseen classes and meanwhile keeping its performance on seen/trained classes from degrading, while the potential problems could be: 1) the output prediction biases towards the seen/training class; and 2) the adapted model suffers from catastrophic forgetting of training classes.
% The learning style is modeled after continual learning and is suitable for problem settings where the model needs to adapt to new problems quickly without forgetting old problems. 

% Zero-shot learning (ZSL)~\cite{awa1} designs a setting to test the model's ``imagination'' on unseen classes.
% The problem is formally defined as if the model has a good generality to the classification on unseen classes, given only the semantic information for every class is available. However, this test setting does not consider the forgetting problem and does not fit for the idea of continual learning, since models may clearly classify within unseen classes but the performance dramatically decreases when considering both the seen and unseen classes. Generalized zero-shot learning (GZSL) is a more practical setting, where the test set includes data from both seen and unseen classes and the prediction is not restricted to the scope of unseen classes. GZSL is an embodiment of continual learning and is suitable for the problem setting where the model needs to quickly adapt to new problems without forgetting the old problems based on the descriptions of the problems even if no training samples are available so far.

The traditional supervised learning cannot work for ZSL due to the lack of training samples for unseen classes. However, even in such an extreme case, it is possible to derive visual patterns for images of unseen classes by exploring the relationship between unseen classes and seen classes, if given their semantic embeddings in the same attribute space, because an unseen class might share attributes with many seen classes.
%the semantic information associated with every class can be used as a proxy for the class labels so that 
Hence, learning a classifier of unseen classes can be reduced to learning a transformation between the attribute space and the visual-feature space~\cite{frome2013devise,xie2019attentive} and a simple distance-based classifier can be generated for unseen classes in the attribute space.
% in turn, the prediction becomes one of distance-based similarity within one of those spaces. 
The method of learning a semantic-image transformation has been the most popular approaches for zero-shot learning during the past several years, and both  linear~\cite{frome2013devise} and nonlinear~\cite{socher2013zero,xian2016latent} transformations have been studied.

In previous works, the attribute space is usually hand-crafted by human experts or pre-trained such as word embeddings of class names, and is independent to the zero-shot learning model. It is the only bridge relating unseen classes and the visual patterns learned for seen classes, so the quality and robustness of ZSL heavily rely on the attribute space. In addition, a pre-defined attribute space cannot fully capture the relationship between different classes which are most important to ZSL. 
In this paper, we optimize the attribute spaces and vectors by training an attribute propagation mechanism across a graph of classes together with the zero-shot classifier model in an end-to-end manner. Consequently, we can refine the attribute vectors of classes to be more informative in ZSL tasks by fully exploring the inter-class relations, which has not been rarely studied in previous works.
% The relationships within the semantic space are a relatively unexplored area, as most current approaches solely focus on the relationships between the semantic and the feature spaces.
%Additionally, most current approaches are based on the assumption every semantic representation is independent of each other. 
The above strategy needs a graph of classes in advance. Fortunately, hierarchical structures and graph of categories are usually free or cheap to acquire in a variety of scenarios, e.g., the species in biology taxonomy, diseases in diagnostic and public heath systems, and merchandise on an e-commerce website.
% it is fairly common to achieve of categories as the number of categories grows. We can see examples of this in the species of biology taxonomy, disease classifications, and merchandise on an e-commerce website. The advantage of category graphs, is that the seen categories/classes and the unseen categories can share the same graph, so regularizing the semantic representations according to their latent and internal structural relationships can benefit the generalizations to unseen classes.
% while previous works have rarely explored the relationship within the space itself and assume every semantic representation is independent to each other. 
% The categories naturally form a hierarchy structure when the number of categories grows, for instance, species in the biology taxonomy, diseases in the classification coding system, and merchandise on an e-commerce website.
% The seen categories/classes and the unseen categories can share the same graph, so that regularizing the semantic representation by their internal latent structured relationship can benefit the generalization over the unseen classes.

In this paper, we aim to optimize the attribute space in the context of ZSL and leverage the inter-class relations to generate more powerful attribute vector per  class. We propose ``attribute propagation network (APNet)'' as a neural nets model propagating the attributes of every class on the category graph to its neighbors in order to generate attribute vectors, followed by a nearest neighbor classifier with learnable similarity metric to produce prediction of an image based on its nearest neighbor among all classes' attribute vectors in the learnable attribute space. The propagation updates attribute vector per class by a weighted sum of the attribute vectors of its neighboring classes on the graph.
% The propagation process passes structural information to every node and improves the generalizability for the unseen nodes/classes. 
When the category graph is unavailable or inaccessible, we further reuse the attention module in propagation to generate a graph of classes given some pre-defined semantic embeddings of the classes. It computes similarity between classes in the semantic space and applies  thresholding to the similarity values in order to determine whether adding an edge between two classes on the graph. 
% To avoid using extra predefined hierarchy/link information, we developed a threshold-based hierarchy generation mechanism based on learned attention.
% The attention learns the similarity between attributes of every class/node pair and the link is predicted between every two classes/nodes if the attention score between the class pair is over a predefined threshold. 
% The propagation follows the built edges/links and assimilates the attribute information of neighboring nodes according to their closeness as indicated in the attention scores. The propagated attributes are, hence, a weighted sum of the attributes from neighbors (including itself).
% The prediction is based on a parametric KNN classifier, which learns a similarity metric between the propagated attribute representation and a query image feature representation.
All the above components can be trained together in an end-to-end manner on training classes and their samples. 
For better generalization on unseen classes and computational efficiency, 
we adopt a meta-learning strategy to train APNet on multiple randomly sampled subgraphs, each defining a classification task over the classes covered by the subgraph. 
% APNet is trained in a meta-learning style strategy on subgraphs, where each graph of classes is a classification task.

% We develop Attributes Propagation Networks (APN) to propagate attributes which are associated with every class on the category graph. 
% Propagation passes structure structure information to every node and improve the generality to the unseen nodes/classes.
% To avoid using extra class hierarchy information (our framework can also integrate with extra hierarchy information), we develop a threshold-based hierarchy generation mechanism based on a learned attention. 
% The attention learns the similarity between every class pair and the edge is connected between every two classes if the attention score between the class pair is over a predefined threshold. 
% The propagation follows the built edges and assimilate the neighbors attributes information according to their closeness defined by the attention scores. 
% The propagted attributes are a weighted sum of the attributes from neighbors (including itself).
% The prediction is transformed into a procedure of attributes retrieval based on a learned similarity metric between the image feature representation and propagated attributes representation.

In experiments on five broadly-used benchmark datasets for ZSL, APNet achieves new state-of-the-art results in the practical generalized zero-shot learning setting and matches the current state-of-the-art results in the zero-shot learning setting. 
We further provide an ablation study of several possible variants of APNet, which show the effectiveness of our propagation mechanism and the improvements caused by the meta-learning training strategy.
% We also compare the performance of some variants of APNet, which shows the flexibility and extendability of our model by integrating with extra hierarchy information.

\begin{figure*}[t!]
%  \vspace{-0.8em}
\begin{center}
\includegraphics[width=2\columnwidth]{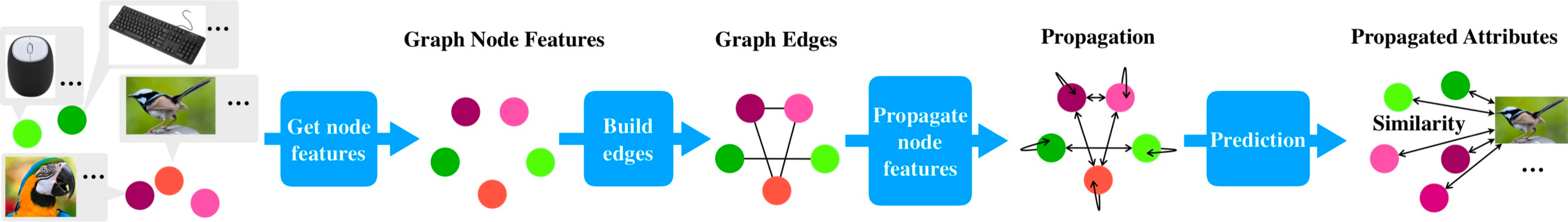}
\end{center}
% \vspace{-1.2em}
\caption{
The pipeline of attribute propagation. \textbf{1. Initialize the nodes features}: the attribute vectors of different classes  are represented by different-colored dots. Each attribute vector is associated with some images from the corresponding class. The features of nodes on the propagation graph are initialized by transforming the attributes using the expert modules from~\cite{zhang2019co}. \textbf{2. Determine the graph edges}: Two nodes on the propagation graph are connected by an edge if the similarity between their feature vectors exceeds a pre-determined threshold. \textbf{3. Propagation on the graph}: The node features are propagated by an attention mechanism. \textbf{4. Zero-shot prediction}: After propagation, a similarity metric is learned between the propagated context-aware/structure-aware attributes representations and a query image feature representation. The class with the largest similarity between the attribute vector after propagation and the query image's embedding in the attribute space are regarded as the predicted class.
% The pipeline. \textbf{1. Get nodes features:} Attributes feature representations are indicated by circles with different colors for different classes. Each attribute representation is associated with images for the corresponding classes. 
% The node features for the propagation graph are obtained by transforming the attributes using the expert modules in ~\cite{zhang2019co}. 
% \textbf{2. Build edges:} Edges of the propagation graph are connected if the similarity of the connected node features is above a predefined threshold.
% \textbf{3. Propagate node features:} The node features are propagated based on attention mechanism using the built edges and the generated node features.
% \textbf{4. Prediction:} After propagation, a similarity metric is learned between the propagated context-aware/structure-aware attributes representations and a query image feature representation. The class with the highest similarity between the propagated attributes and the query image representation is regarded as the prediction result.
}
\label{fig:spn}
% \vspace{-1.5em}
\end{figure*}

\section{Related Work}
%The aim of zero-shot learning is to learn a model that is generalizable to new classes or tasks that are described but where no training data is available
Zero-shot learning (ZSL)~\cite{Larochelle:2008:ZLN:1620163.1620172,lampert2013attribute,xian2018zero} is an open problem that has been studied for a long time.
Traditional methods learn a transformation between the semantic attribute space and visual-feature space (e.g., a hidden space of deep neural nets).  
In these works, given a learned similarity metric, ZSL is cast into a retrieval task, where the label is determined by retrieving the associated attributes from a 
% The similarity can be defined as the distance between the query feature vector and the 
set of candidate vectors based on the learned transformation~\cite{frome2013devise,akata2015label,akata2015evaluation,romera2015embarrassingly,kodirov2017semantic,xian2016latent,socher2013zero}.
% They cast the problem as retrieval where the assigned label is associated with the feature vector with the minimum distance between the query feature vector based on the learned transformation space~\cite{frome2013devise,akata2015label,akata2015evaluation,romera2015embarrassingly,kodirov2017semantic,xian2016latent,socher2013zero}.
An implicit relationship between seen and unseen classes is that the properties of the unseen classes can be regarded as a mixture of properties of the seen classes. For example, a mixture of the semantic features of the seen classes~\cite{norouzi2013zero} or a mixture of the weights of some phantom classes trained on the seen classes~\cite{changpinyo2016synthesized}. 

% A more indirect relationship between seen classes and unseen classes is that the properties of the unseen classes can also be regarded as a mixture of properties for seen classes, e.g. a mixture of the semantic features of seen classes~\cite{norouzi2013zero}, a mixture of the weight of some phantom classes trained on the seen classes~\cite{changpinyo2016synthesized}.
% Our method does not have an explicit transformation function between different modalities. We achieve transformation by cross-propagation over class-level, i.e., propagation between visual prototypes and semantic prototypes within one class and between classes following paths defined on category graph.

More recently, researchers start to integrate knowledge graph with zero-shot recognition~\cite{wang2018zero,kampffmeyer2019rethinking}.
% \cite{wang2018zero} proposed a scheme to predict the weight of a fully-connected classifier based on the corresponding semantic feature vectors by using a graph neural network. 
% The node features are the semantic feature vectors for each class and the output is the weight of the fully-connected classifier for that class.
% \cite{kampffmeyer2019rethinking} extended the above idea: it adopts a densely-connected graph neural network and add extra fine-tuning of the backbone CNN. Our method, APNet, learns a propagation mechanism over an attribute space to generate an attribute vector per class,
% %which then is compared to the image features for the similarity calculation.
% which is then compared with images' embeddings in the same space to produce class prediction results.
They use graph convolutional network (GCN) which relies on a given knowledge graph to provide adjacency matrix and edges. In contrast, we use a modified graph attention network (GAN)~\cite{DBLP:conf/aaai/ShenZLJPZ18,DBLP:conf/naacl/ShenZL0Z19},  which is capable to learn both the similarity metric and edges by itself even without any pre-defined graph or metric. 
Thereby, the propagation scheme learned in APNet is more powerful in modeling intra-task relationships and can be applied to more practical scenarios. For this reason, they cannot be applied to most ZSL benchmark datasets, which do not provide graphs.
Their output of GCN is a fully connected layer per class, as a single-class classifier, that aims to approximate the corresponding part of the last layer of a pre-trained CNN, while our output of APNet is a prototype per class, from which we build a KNN classifier for each task. Hence, they need the CNN to be able to predict all the possible classes for all few-shot tasks. In contrast, we do not require any ground truth for the per-class classifier/prototype, so any pre-trained CNN can be used to provide features for APNet.
% Moreover, generating FC layer weights from scratch in their methods turns to be a harder problem than refining prototypes from a reasonably good initialization by propagation in our method.

% Recent works utilize knowledge graph propagation for zero-shot recognition.~\cite{wang2018zero} proposed to predict the weight of a fully-connected classifier weight by the corresponding semantic feature vectors. They achieved this by a graph neural networks where the node feature is the semantic feature vector for each class and the output is the weight for the fully-connected classifier for this class.
% ~\cite{kampffmeyer2019rethinking} extended the work by a densely-connected graph neural network and an extra fine-tuning for the CNN afterwards.
% % However, our work aims to learn the feature prototype for every class instead of model weight and applies KNN classifier based on the learned prototypes.  
% Our method learns a propagation mechanism over the 
% semantic space to generate a context-aware representation, which then is compared to the image features for similarity calculation.

Our idea of attribute propagation is inspired by belief propagation, message passing and label propagation.
It is also related to Graph Neural Networks (GNN)~\cite{henaff2015deep,wu2019comprehensive}, where convolution and attention are iteratively applied over a graph to construct node embeddings. 
In contrast to our work, their task is defined on graph-structured data, i.e., node classification~\cite{hamilton2017inductive}, graph embedding~\cite{pan2018adversarially}, and graph generation~\cite{dai2018syntax}. We only use the relationships between the categories/classes and build a computational graph that passes messages along the graph hierarchy. Our training strategy is inspired from the meta-learning training strategy proposed in~\cite{santoro2016meta}, which have been broadly used in the meta-learning and few-shot learning literature~\cite{finn2017model,snell2017prototypical,dong2019search}, especially graph meta-learning~\cite{liu2019ppn,liu2019GPN}. However, this paper addresses zero-shot learning problem by training a novel model, i.e.,  APNet.

\section{Problem Formulation}
Zero-shot learning aims to learn a model that is generalizable to new classes or tasks whose semantic attributes are given but no training data is provided. For example, the model is expected to be applied to a classification task over some seen and unseen classes, 
%but the unseen classes have no data, only semantic descriptions. 
The semantic attributes could have the form of an attribute vector, such as the main color of the class, or a word embedding  of the class names.

Formally, we assume that a training set $\mathcal{X}^{tr}$ and a test set $\mathcal{X}^{te}$ are sampled from a data space $\mathcal X$. Each training data $\vx \in \mathcal{X}^{tr}$
is annotated with a label $y \in \mathcal{Y}^{seen}$. 
The model is tested over $\mathcal{X}^{te}$, which are not only from the seen classes $\mathcal{Y}^{seen}$ but also the unseen classes $\mathcal{Y}^{unseen}$.
The challenge is that the seen and unseen classes have no overlaps: $\mathcal{Y}^{seen} \cap \mathcal{Y}^{unseen} = \emptyset$.
Hence, the semantic attributes $\mS$ for each class is made available during both training and testing to act as a bridge between training classes and test classes. Specifically, every class $y \in \mathcal{Y}^{seen} \cup \mathcal{Y}^{unseen} $ is associated with a semantic embedding vector $\vs_y \in \mS$.

% Formally, each training data $\vx^{tr} \in \mathcal{X}^{tr}$ is annotated by a label $y \in \mathcal{Y}^{seen}$. The model is tested over a set of data $\mathcal{X}^{te}$, which includes data from not only the seen classes $\mathcal{Y}^{seen}$ but also unseen classes $\mathcal{Y}^{unseen}$. The challenge is that the seen classes and the unseen classes have no overlap between each other: $\mathcal{Y}^{seen} \cap \mathcal{Y}^{unseen} = \emptyset $.
% To alleviate this challenge, semantic information $\mS$ for each class is also available during training and test so that the training classes and test classes can be connected by their semantic relationships. Specifically, every class $y \in \mathcal{Y}^{seen} \cup \mathcal{Y}^{unseen} $ is associated with one semantic vector $\vs \in \mS$.

A direct mapping $F:\mathcal X \mapsto \mathcal{Y}$ from data to label is difficult to learn with zero-shot learning because the training and testing data are non-i.i.d. An alternative is to learn a mapping from data to the semantic attributes, i.e., $F:\mathcal{X}\mapsto\mS$. In APNet, we learn a parameterized KNN classifier: the model learns a metric to measure the similarity between class-level semantic attributes and image representations. Here, the semantic attribute vector is a descriptor to the class label. Moreover, the higher the similarity, the higher the probability that the image belongs to the corresponding class. The hierarchical relationships between classes are incorporated into the semantic representations through an attribute propagation mechanism that traverses a graph. In the following sections, we will discuss how to build the propagation graph, how to propagate along it, and how to make predictions based on the propagated attribute representations.

\section{Attribute Propagation Network}
\label{model}

We propose the attribute propagation network (APNet) for zero-shot learning. APNet propagates attribute representations of each class between each other based on a semantic distance or class hierarchy, shared by both the seen classes and unseen classes, so that the propagated attribute representations can leverage the latent relationships between classes and generate context-aware/hierarchy-aware attribute representations. We learn a parametric KNN classifier since KNN classifier has better generality when the data is limited.
The prediction in KNN depends on a learned similarity metric used to measure the discrepancy between the query image and the propagated semantic representation for each candidate class. 
The learned similarity is easier to generalize to the unseen classes based on the graph structure-encoded attribute representations.
However, challenges still remain. In the rest of this section, we illustrate solutions to four of these challenges. 
1. How to choose the nodes for the propagation graph? 
2. How to represent the node features in the propagation graph? 
3. How to find the edges in the propagation graph? 
4. What is the best propagation mechanism?

% We propose Attributes Propagation Networks (APN) for zero-shot learning.
% APN propagates attributes representation of each class between each other based on their semantic distances/ class hierarchy.
% In this way, the propagated attributes representation can utilize the latent relationships between classes and generate context-aware/class-hierarchy-aware attribute representations.
% The challenge of the classification problem over both seen and unseen classes is mitigated by casting the problem into a regression problem of the similarity between and the attributes and the image representations.
% To improve the generality of the regression model, we learn to propagate the attributes so that the graph structure information, shared by both the seen classes and the unseen classes, are integrated in the propagated attributes.
% The prediction is based on a learned similarity metric between the query image feature and the propagated semantic representation for each candidate classes.
% The learned similarity is easier to generalize to the unseen classes based on the graph structure encoded attribute representations. 
% The rest of the questions are: 1. How to choose the nodes of the propagation graph? 2. How to get the node features of the propagation graph? 3. How to get the edges of the propagation graph? 4. What is the propagation mechanism?

\subsection{Choosing the Graph Nodes}

We assume propagation is built on a graph $\gG=(\mathcal Y, \mathcal E)$, where each node $y\in \mathcal Y$ denotes a class, and each edge (or arc) $(y,z)\in\mathcal E$ connects two classes $y$ and $z$ and serves as a propagation pathway. Previous works built their propagation graph from a predefined class hierarchy~\cite{wang2018zero,kampffmeyer2019rethinking}, e.g., the WordNet class hierarchy for ImageNet~\cite{imagenet}. In their approaches, each directed edge (or arc) $y\rightarrow z\in \mathcal E$ connects a parent class $y\in\mathcal Y$ to one of its child classes $z\in\mathcal Y$ on the graph $\gG$. However, this kind of graph construction has some limitations: 1) It needs extra hierarchy information, which may not always be available; 2) The graph’s structure needs to be intact, and there needs to be a subgraph that covers all the targeted classification classes/nodes as well as their neighbors. A sparse graph containing neighbor nodes with many missing features would introduce noise when passing information through connected nodes, which might lead to a computational overhead that prevents the propagation being completed within a few steps.
%especially since the computational overhead which means that propagation must be completed within a few steps. 
To flexibly apply graph propagation to classes regardless of how they are distributed in the actual class hierarchy, nodes with missing information are excluded from the propagation stage. In this way, the propagation process is not disturbed by those ``blank'' nodes uncovered by the training set.

\subsection{Node Feature Representations}
In this work, we assume that each node is associated with a feature representation. Encoded attributes for each class/node $y\in\mathcal{Y}$ are used for the initial node feature representations $\mX_{y}^{0}$, following the model in~\cite{zhang2019co}:
% \begin{align}\label{equ:preprocess}
% \vx = \sum_{i=1}^{\|\mC\|} ReLU{i}(g_{i}(\vs-\mC_{i})), \\
% \mP^{s}_{y}[0] = \sum_{i=1}^{k} \relu[\Theta_{i}(\vs_y-C_{i})],~~\forall~y\in\mathcal Y^T,
% \end{align}
\begin{equation}\label{equ:preprocess}
\mX^{0}_{y}= \sum_{i=1}^{k} \relu[\Theta_{i}(\vs_y-C_{i})],
\end{equation}
where $\vs_{y} \in \mS$ is the provided side information, i.e., the class name embeddings for every class/node. $\mC$ is the matrix for the centroids of the class attribute space, which is generated by $k$-means clustering and covers all training class attributes vectors $\mS$. Each line $\mC_{i}$ in the matrix stores one of the centroids. $\Theta_i$ is a linear transformation. We chose this model for its  efficiency, since multiple attributes can share the encoding parameters $g_i$. Therefore, the number of parameters only grows linearly according to the number of centroids $|\mC|$. Less learnable parameters can reduce problems with overfitting, which are common with zero-shot learning. Also, after encoding, the semantic information with different dimensions is unified so that the subsequent modules can use parameters of the same dimensions.

% In this work, we assume each node is associated with a feature representation $\vx \in \mX$. We use encoded attributes for each class/node for the node features representations following the module in~\cite{zhang2019co}:
% \begin{equation}\label{equ:preprocess}
% \vx = \sum_{i=1}^{\|\mC\|} \sigma_{i}(g_{i}(\vs-\mC_{i})),
% \end{equation}
% where $\vs \in \mS$ is the provided side information, i.e., class attributes for every class/node, 
% $\mC$ is the matrix for the centroids of the class attributes space covering all training class attributes vectors $\mS$, where each line $\mC_{i}$ stores one of the centroids, $g_i$ is a learnable transformation, e.g., linear transformation, for normalization based on a centroid $\mC_{i}$ and $\sigma_{i}$ is the nonlinear function. 
% We choose this module due to the efficiency, since multiple attributes can share the encoding parameters $g_{i}$ so that the number of parameters only grows linearly according to the number of centroids $\|\mC\|$.
% Less learnable parameters can alleviate the overfitting problem which usually occured in zero-shot learning.

% This can be regarded as an ensembled version of multiple normalizations based on multiple centroids. The technique of ensemble is to stabilize the normalization for data with different distributions and improve the generality of the normalization, especially when new/unseen semantic information is included in the test stage. This is also beneficial to continual learning when new classes are sequentially included in the task.

\subsection{Finding the Graph Edges}
As mentioned, the graph nodes are selected as a batch of nodes/classes that are available in the current task, with encoded class attributes as the node feature representation. Now we need to connect the nodes by deriving an adjacency matrix of the graph. An intuitive way to do this is to define a value in the adjacency matrix between node $i$ and node $j$ according to the predefined distance in the class hierarchy of the whole graph. For example, for the classes in ImageNet~\cite{imagenet}, 
each entry in adjacency matrix would be $A_{ij} = 1/d_{ij}$, where the distance $d_{ij}$ is the number of hops between node $i$ and node $j$ on WordNet. 
(Note that we tried this type of adjacency matrix, and it generated similar results to APNet). Unfortunately, this kind of strategy requires extra information about the class hierarchy, e.g., WordNet. Hence, in our approach, we propose to generate the graph edges according to the node feature representations using an attention mechanism. More specifically, the attention learns a similarity metric $a(\cdot, \cdot)$ between node feature representation pairs and generates an edge between pairs with high similarity. The set of edges on graph $\mathcal G$ is:
% \begin{align}
% A_{ij} = \mathbbm{1}[a(\vx_{i}, \vx_{j}) > \epsilon], \label{equ:adj-generation} \\
% a(p, q)\triangleq\frac{\langle f(p), f(q)\rangle}{\|f(p)\| \times \|f(q)\|}, \label{equ:attention}
% \end{align}
\begin{align}\label{equ:adj-generation}
\mathcal E = \left\{(y,z): y,z\in\mathcal Y, a(\mX_{y}^{0}, \mX_{z}^{0})\geq\epsilon\right\}, \\
a(p, q)\triangleq\frac{\langle f(p), f(q)\rangle}{\|f(p)\|_2 \times \|f(q)\|_2}, \label{equ:attention}
\end{align}

where $f(\cdot)$ is a learnable transformation; and $\epsilon$ is a threshold for the similarity of the edge connection. For, simplicity, the edges are assumed to be undirected.

\begin{figure}[t!]
% \vspace{-0.8em}
\begin{center}
\includegraphics[width=\columnwidth]{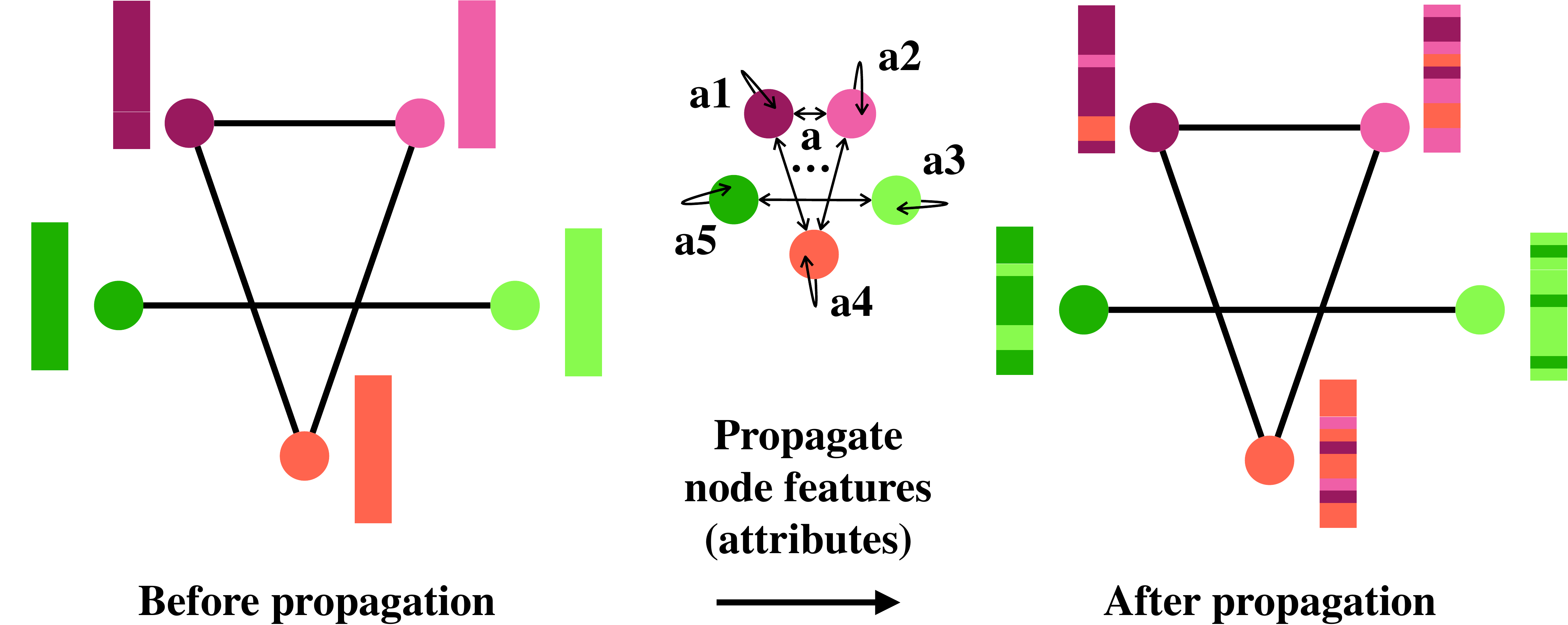}
\end{center}
% \vspace{-1.0em}
\caption{
One-step of attribute propagation. \textbf{Before propagation}: The nodes, denoted by different-colored circles represent the different classes and the node features are the corresponding attribute feature representation of this class, denoted in the columns. The edges are generated based on the similarity between every node feature pair. \textbf{Propagating the node features (attributes)}: Propagation computes a weighted sum of its neighbors’ node features, where the weights are produced by an attention module applied to every node feature and its neighbors (including itself). \textbf{After-propagation}: The attribute representations comprise a mixture of attribute representations from neighbors.
% One step attributes propagation. \textbf{Before propagation:} The nodes, indicated by circles with different colors, represent different classes and the node features are the corresponding attribute feature representation of this class, indicated by columns. The edges are generated based on the the similarity between every node features pair.
% \textbf{Propagate node features (attributes)} Propagation is based on an attention mechanism: propagated node features are a weighted sum of the neighbors' node features, where the weight is based on the attention score between every node feature and its neighbor features (including itself).
% \textbf{After propagation:} The attributes representations integrate a mixture of attribute representations from neighbors.
}
\label{fig:prop}
% \vspace{-1.0em}
\end{figure}

\subsection{Attribute Propagation}
The attribute propagation is based on the graph constructed in the previous subsections. The propagation is bidirectional through the edges using the learned attention mechanism. The node feature representations are a weighted sum of the features representations of its neighboring nodes $\mathcal N_{y}$, whose weights were obtained from the attention
mechanism using~\Eqref{equ:attention}. 
At step t, the propagation proceeds as follows:
\begin{align}\label{equ:prop}
    \mX^{t+1}_{
    y}\leftarrow\sum_{z\in\mathcal N_{y}} a'(\mX^{t}_{y},
    \mX^{t}_{z}) \times \mX^{t}_{z},
\end{align}
where $a'(\cdot,\cdot)$ is the normalized attention score over the neighbors $\mathcal N_{y}$ using softmax with hyper-parameter temperature $\gamma$ that controls the smoothness of the softmax function: 
\begin{align}\label{equ:normalize-att}
   a'(\mX^{t}_{y}, \mX^{t}_{z}) = \frac{ \exp[\gamma_{1} a(\mX^{t}_{y}, \mX^{t}_{z})] }{ \sum_{z\in\mathcal N_{y}} \exp[\gamma_{1} a(\mX^{t}_{y}, \mX^{t}_{z})]}.
\end{align}
The propagation can be applied multiple times to collect messages from indirectly-connected neighbors and, in so doing, amass a more comprehensive understanding of the graph structure. Figure~\ref{fig:prop} shows an example of one-step propagation.

\subsection{Parameterized KNN for Prediction}
In this approach, we learn a similarity metric between an attribute representation and a query image representation as the parametric KNN. 
A learned metric calculates the similarity between the $t$-th step propagated attribute matrix $\mX^{t+1}$ and the query image feature vectors. Inspired by the concept of additive attention ~\cite{vaswani2017attention}, the similarity between an image with feature $\vx$ and class $y$ is:
\begin{align}\label{equ:similarity-pred}
  h(\mX^{t+1}_{y}, \vx) =  \vw \sigma^{(1)}(\mW^{(1)} \mX^{t+1}_{y} + \mW^{(2)} \vx + \vb^{(1)}) + b,
\end{align}
where $\mW^{(1)}$, $\mW^{(2)}$, $\vb$, $\vw$ and $b^{(1)}$ are learnable parameters, and $\sigma^{(1)}$ is a nonlinear activation function.
The class whose propagated  attribute vector with the highest similarity to the image vector is taken as the final prediction. The probability of predicting $\vx$ as class $y$ normalized with temperature $\gamma_{2}$ over a batch of classes $\mathcal Y^T$ is:
% The optimal similarity is binary and defined as if or not the label of an image  is the same as the label associated with an attribute representation:
\begin{align}\label{equ:similarity-label}
\Pr(y|\vx;\mX^{t+1}) = \frac{\exp(\gamma_{2} h(\mX^{t+1}_{y}, \vx))}{\sum_{z\in \mathcal Y^T}\exp(\gamma_{2} h(\mX^{t+1}_{z}, \vx))}.
% h^{*}(\vx^{j},\vx^{tr}) = \mathbbm{1}[y^{\vx^{j}} = y^{\vx^{tr}}],
\end{align}
% where $y^{\vx^{j}}$ and $y^{\vx^{tr}}$  indicate the label associated with attribute representation $\vx^{j}$ and the label of image representation $\vx^{tr}$. 
We assume that the attribute propagation does not affect the associated attribute labels so the label for the propagated attributes is the same as the original associated label before propagation, i.e., $\mX^{t}_y$ is the attribute for class $y$ before propagation while the $\mX^{t+1}_y$ is still the attribute for class $y$ after propagation.

\subsection{Training Strategy and Scalability}
Training APNet over all the nodes/classes in a dataset would be too computationally expensive since the size of the attention score matrix grows exponentially as the number of nodes/classes grows. Moreover, applying a cost constraint over the loss computation may make the optimization unstable. Hence, for training efficiency, we only sample a subgraph to apply the propagation per iteration. Additionally, to improve the generality of our model to unseen nodes/classes, we train APNet on different classification tasks in every iteration, so it can quickly adapt to different new tasks. This technique was inspired by the training strategy of meta-learning originally developed for few-shot learning in~\cite{vinyals2016matching}. The aim of few-shot learning is to build a model that can adapt quickly to any new task with only a few training samples. In contrast, the goal of zero-shot learning is to learn a model that can generalize to a new task with only the semantic information provided. However, we found the idea of meta-learning training strategy designed for “fast adaptability” could also improve the generality of our model. Interestingly, this training strategy might also benefit other models in the literature of zero-shot learning.\looseness-1

% Training APN over all nodes/classes in the dataset is too computationally expensive: the size of the attention score matrix grows exponentially as the number of nodes/classes grows. 
% Applying a constrain over the the computation cost in the loss may make the optimization unstable. 
% In order to efficiently train APN, we do propagation on a subgraph in each iteration.
% Besides, to improve the generality of our model, especially the generality on the unseen nodes/classes, we train APN over different classification tasks in every iteration, i.e., classify different sets of classes. This is inspired by the ``episode training'' strategy, which is originally proposed for few-shot learning in~\cite{vinyals2016matching}. 
% Few-shot learning aims to learn a model with fast adaptability on any new task with only a few training samples while the goal of generalized zero-shot learning is to learn a model which can generalize to a new task with only semantic information provided.
% We found the idea of ``episode training'' designed for ``fast adaptability''can also improve the ``generality'' of our model. This training strategy can also benefit other models in the literature of zero-shot learning.

In every iteration, a task $T$ is sampled from a task distribution $\gT$. The aim is to minimize the loss for this classification task based on our APNet model.
The training objective is to minimize the empirical risk:
% The training objective is to minimize the mean squared error between the similarity prediction and the optimal similarity using \Eqref{equ:similarity-pred} and \Eqref{equ:similarity-label}, respectively:
\begin{align}\label{equ:opt-obj-all}
\min_\Theta{\sE}_{T\sim\gT}{\sE}_{(\vx, y)\sim\gD^{T}} -\log\Pr(y|\vx;\mX^{T}),
% (h(\vx^{j},\vx^{tr}) - h^{*}(\vx^{j},\vx^{tr}))^{2},
\end{align}
where each task $T$ is defined as a subset of classes $\mathcal Y^T\subseteq \mathcal Y^{seen}$; $\mathcal D^{T}$ is the distribution of data-label pair $(\vx,y)$ with $y\in \mathcal Y^{seen}$; $\mX^{T}$ is the corresponding propagated attributes matrix for classes in task $T$; $\Theta$ are the learnable parameters.

% In every iteration, a task $T$ is sampled from a task distribution $\gT$, we aim to minimize the loss for this classification task based on our APN model introduced previous subsections.
% The training objective is to minimize the mean squared error between the similarity prediction and the optimal similarity using \Eqref{equ:similarity-pred} and \Eqref{equ:similarity-label}, respectively:
% \begin{align}\label{equ:opt-obj-all}
% %\min_\Theta\mathbb E_{T\sim\mathcal T}\left[\mathbb E_{(x,y)
% %\sim\mathcal D^T}  [ \mathbb E_{\vs^{0} \in \mS^{T}} (\hat{a}-a)^{2} ]\right],
% %\min_\Theta\mathbb E_{T\sim\mathcal T}\left[\mathbb E_{(x,y)\sim\mathcal D^T, \vs^{0} \in \mS^{T}}  (\hat{a}-a)^{2} \right], \\
% \min_\Theta{\sE}_{T\sim\gT}{\sE}_{\vx^{tr}\sim\gD^{T}, \vx^{j} \in \mX^{T}}  (h(\vx^{j},\vx^{tr}) - h^{*}(\vx^{j},\vx^{tr}))^{2},
% %\hat{a} = \hat{a}(\vs^{t},\vx|\Theta, \vs^{0}), a = a(\vs^{t},\vx | \vs^{0}),
% \end{align}
%\begin{equation}\label{equ:opt-obj-split}
%\hat{a} = \hat{a}(\vs^{t+1},\vx|\Theta, \vs^{0}), a = a(\vs^{t+1},\vx | \vs^{0})
%\end{equation}
% \noindent where each task $T$ is defined by a subset of classes $\mathcal Y^T\subseteq \mathcal Y^{seen}$; $\mathcal D^{T}$ is the distribution of data-label pair $(\vx^{tr},y)$ with $y\in \mathcal Y^{seen}$; $\mX^{T}$ is the corresponding propagated attributes matrix for classes in task $T$; $\Theta$ is the learnable parameters.

\begin{table*}[t!]
\centering
% \vspace{-1em}
% \setlength{\tabcolsep}{4pt}
% \vspace{-1em}
\caption{
Datasets Statistics. ``\#*'' denotes the number of *. “Tr-S”, “Te-S” and “Te-U” denotes seen classes in training, seen classes in test and unseen classes in test, respectively.
}
\resizebox{2\columnwidth}{!}{
\begin{tabular}{lccccccc}
\toprule
\textbf{Dataset}     & \textbf{Granularity} & \textbf{\#Attributes} & \textbf{\#Seen classes} & \textbf{\#Unseen classes} & \textbf{\#Imgs(Tr-S)}  & \textbf{\#Imgs(Te-S)} & \textbf{\#Imgs(Te-U)} \\ \hline
SUN  & fine & 102 & 645 & 72 & 10,320 & 2,580 & 1,440 \\
CUB  & fine & 312 & 150 & 50 & 7,057  & 1,764 & 2,967 \\
AWA1 & coarse & 85 & 40 & 10 & 19,832 &  4,958 & 5,685 \\
AWA2 & coarse & 85 & 40 & 10 & 23,527 &5,882 & 7,913 \\
aPY & coarse & 64 & 20 & 12 & 5,932 &  1,483 & 7,924 \\
\bottomrule
\end{tabular}}
% \vspace{-1em}
\label{table:datasets}
\end{table*}

\begin{table*}[t!]
\centering
\caption{
Performance comparisons for generalized zero-shot learning between our method and baselines on five datasets, where ``S'' denotes per-class accuracy  (\%) for seen classes, ``U'' denotes per-class accuracy (\%) for unseen classes and ``H'' denotes harmonic mean of ``S'' and ``U''. Parts of the results are from~\cite{xian2018zero}.
}
% \vspace{2mm}
\resizebox{2\columnwidth}{!}{
\begin{tabular}{l|ccc|ccc|ccc|ccc|ccc}
\toprule
\multirow{2}{*}{\textbf{Methods}}    & \multicolumn{3}{c|}{\textbf{SUN}}  & \multicolumn{3}{|c|}{\textbf{CUB}} & \multicolumn{3}{|c|}{\textbf{AWA1}} & \multicolumn{3}{|c}{\textbf{AWA2}} & \multicolumn{3}{|c}{\textbf{aPY}} \\ \cline{2-16}
 & \textbf{S} & \textbf{U} & \textbf{H}  & \textbf{S} & \textbf{U} & \textbf{H}  & \textbf{S} & \textbf{U} & \textbf{H}  & \textbf{S} & \textbf{U} & \textbf{H} & \textbf{S} & \textbf{U} & \textbf{H} \\
\midrule
DEVISE~\cite{frome2013devise}  & 27.4 & 16.9 & 20.9 & 53.0 & 23.8 & 32.8 & 68.7 & 13.4 & 22.4 & 74.7 & 17.1 & 27.8 & 76.9 & 4.9 & 9.2  \\
CONSE~\cite{norouzi2013zero} & 39.9 & 6.8 & 11.6 & \textbf{72.2} & 1.6 & 3.1 & 88.6 & 0.4 & 0.8 & 90.6 & 0.5 & 1.0 & \textbf{91.2} & 0.0 & 0.0 \\
% CMT~\cite{socher2013zero} & 21.8 & 8.1 & 11.8 & 49.8 & 7.2 & 12.6  & 87.6  & 0.9 & 1.8 & 90.0 & 0.5 & 1.0 & 85.2 & 1.4 & 2.8 \\
% ALE~\cite{akata2015label} & 33.1 & 21.8 & 26.3 & 62.8 & 23.7 & 34.4 & 76.1 & 16.8 & 27.5 & 81.8 & 14.0 & 23.9 & 73.7 & 4.6 & 8.7 \\
% IAP~\cite{awa1} & 37.8 & 1.0  & 1.8 & 72.8 & 0.2 & 0.4 & 78.2 & 2.1 & 4.1 & 87.6 & 0.9 & 1.8 \\
SYNC~\cite{changpinyo2016synthesized} & \textbf{43.3} & 7.9 & 13.4 & 70.9 & 11.5 & 19.8 & 87.3 & 8.9 & 16.2 & 90.5 & 10.00 & 18.0 & 66.3 & 7.4 & 13.3 \\
SAE~\cite{kodirov2017semantic} & 18.0 & 8.8 & 11.8 & 54.0 & 7.8 & 13.6 & 77.1 & 1.8 & 3.5 & 82.2 & 1.1 & 2.2 & 80.9 & 0.4 & 0.9 \\
DEM~\cite{zhang2017learning} & 34.3 & 20.5 & 25.6 & 57.9 & 19.6 & 29.2 & 84.7 & 32.8 & 47.3 & 86.4 & 30.5 & 45.1 & 75.1 & 11.1 & 19.4 \\
RN~\cite{sung2018learning} & - & - & - & 61.1 & 38.1 & 47.0 & \textbf{91.3} & 31.4 & 56.7 & \textbf{93.4} & 30.0 & 45.3 & - & - & - \\
% CVAE~\cite{mishra2018generative} & - & - & 26.7 & - & - & 34.5 & - & - & 47.2 & - & - & 51.2 & - & - & - \\
% GAFE~\cite{ijcai2019-graph-auto} & 31.9 & 19.6 & 24.3 & 52.1 & 22.5 & 31.4 & 76.6 & 25.5 & 38.2 & 78.3 & 26.8 & 40.0 & 68.1 & 15.8 & 25.7 \\
% GDAN~\cite{huang2018generative} & 89.9 & 38.1 & 53.4 & 66.7 & 39.3 & 49.5 & - & - & - & 67.5 & 32.1 & 43.5 & 75.0 & 30.4 & 43.4 \\
% GMN~\cite{sariyildiz2019gradient} & 33.0 & 53.2 & 40.7 & 54.3 & 56.1 & 55.2 & 71.8 & 61.1 & 65.8 & - & - & - & - & - & - \\
% VSE~\cite{zhu2019generalized} & - & - & - & 68.9 & 39.5 & 50.2 & - & - & - & 88.7 & 45.6 & 60.2 & 78.7 & 43.6 & 56.2 \\
PQZSL~\cite{Li2019CompressingUI} & 35.3 & 35.1 & 35.2 & 51.4 & 43.2 & 46.9 & 70.9 & 31.7 & 43.8 & - & - & - & 64.1 & 27.9 & 38.8 \\
% MLSE~\cite{ding2019marginalized} & 36.4 & 20.7 & 26.4 & 71.6 & 22.3 & 34.0 & - & - & - & 83.2 & 23.8 & 37.0 & 74.3 & 12.7 & 21.7\\
CRNet~\cite{zhang2019co} & 36.5 & 34.1 & 35.3 &
56.8 & 45.5 & 50.5 & 74.7 & 58.1 & 65.4 & 78.8 & 52.6 & 63.1 & 68.4 & 32.4 & 44.0 \\
\midrule
APNet(ours) & 40.6 & \textbf{35.4} & \textbf{37.8} & 55.9 & \textbf{48.1} & \textbf{51.7} & 76.6 & \textbf{59.7} & \textbf{67.1} & 83.9 & \textbf{54.8} & \textbf{66.4} & 74.7 & \textbf{32.7} & \textbf{45.5} \\
\bottomrule
\end{tabular}
}
% \vspace{-2mm}
\label{table:results-gzs}
\end{table*}

\section{Experiments}

\subsection{Datasets}\label{sec:dataset}
We used five widely-used zero-shot learning datasets in our experiments: AWA1~\cite{lampert2013attribute}, AWA2~\cite{xian2018zero}, SUN~\cite{sun-database}, CUB~\cite{welinder2010caltech} and aPY~\cite{farhadi2009describing}.
To avoid overlaps between the test sets and ImageNet-1K, which is used for pretraining backbones, we followed the splits proposed in~\cite{xian2018zero}. AWA1, AWA2 and CUB are subsets of ImageNet.
AWA1 and AWA2 have 50 animals classes with pre-extracted feature representations for each image. CUB is also a subset of ImageNet with images of 200 bird species (mostly North American). SUB is a Scene benchmark containing 397 scene categories. aPY is a small dataset including 32 classes. Detailed statistics are provided in \Tabref{table:datasets}.

% We use five widely used zero-shot learning datasets in our experiments: AWA1~\cite{lampert2013attribute}, AWA2~\cite{xian2018zero}, SUN~\cite{sun-database}, CUB~\cite{welinder2010caltech} and aPY~\cite{farhadi2009describing}. To avoid overlapping between the test sets and the ImageNet-1k, which is used for pretraining backbones, we follow the split proposed in~\cite{xian2018zero}.
% AWA1, AWA2 and CUB are subsets of ImageNet. AWA1 and AWA2 both have 50 animals classes with pre-extracted feature representations for each image. CUB is an image dataset with photos of 200 bird species (mostly North American). SUN is a Scene benchmark containing 397 scene categories. The detailed statistics is shown in Table.~\ref{table:datasets}.
% aPY is a small dataset including 32 classes.

\subsection{Implementation Details}
For a fair comparison with baselines, we followed~\cite{xian2018zero}'s implementation and used a pre-trained ResNet-101~\cite{he2016deep} on ImageNet-1K to extract 2048-dimensional image features with no fine-tuning on the backbone. We trained our APNet with Adam~\cite{kingma2015adam} for 360 epochs with weight decay factor of 0.0001. The initial learning rate was 0.00002 with a decrease of 0.1 every 240 epochs. The number of iterations in every epoch under an $N$-way-$K$-shot training strategy was $\|\mathcal{X}^{tr}\| / NK$, where $N$ was 30 and $K$ was 1 in our exeperiments.
The temperature $\gamma_{1}$ was 10 and $\gamma_{2}$ was 30.
Transformation functions $g_{i}$ and $f$ were linear transformations.
The threshold for connecting edges was set to $\textrm{cosine}40^{o}~\approx~0.76$. All nonlinear functions were ReLU except for $\sigma$, which was implemented using Sigmoid to map the result between 0 and 1.

% To make a fair comparison with baselines, we follow the implementation by~\cite{xian2018zero} to use the pretrained ResNet-101~\cite{he2016deep} on ImageNet-1K to extract the 2048-dimensional image features with no fine-tuning on the backbone.
% Our APN is trained using Adam~\cite{kingma2015adam} for 360 epochs with weight decay factor of 0.0001.
% The initial learning rate is 0.00002 and is decreased by 0.1 after 240 epochs.
% To keep the steps for backpropagation roughly the same with traditional minibatch training when the number of epochs the same, the number of iterations in every epoch under a $N$-way-$K$-shot training strategy is $\|\mathcal{X}^{tr}\| / NK$, where $N$ is 30 and $K$ is 1 in our exeperiments.
% The temperature $\gamma$ for normalizing is 10.
% Transformation functions $g_{i}$ and $f$ are linear transformations.
% The threshold for connecting edges is set to $\textrm{cosine}(40^{o})~\approx~0.76$ so that the connected nodes should be at least around $3/4$ similar to each other compared to the similarity level within the node itself. All nonlinear functions are ReLU except for $\sigma$. $\sigma$ is implementated using Sigmoid to map the result between 0 and 1.

\begin{table*}[t!]
\centering
\caption{Per-class accuracy (\%) comparisons over unseen classes for zero-shot learning between our method and baselines on five datasets. Parts of the results are from~\cite{xian2018zero}.
}
\begin{tabular}{l|c|c|c|c|c}
\toprule
\textbf{Methods}    & \textbf{SUN}  & \textbf{CUB} & \textbf{AWA1} & \textbf{AWA2} & \textbf{aPY}\\ 
%  & \textbf{S} & \textbf{U} & \textbf{H}  & \textbf{S} & \textbf{U} & \textbf{H}  & \textbf{S} & \textbf{U} & \textbf{H}  & \textbf{S} & \textbf{U} & \textbf{H} \\
\midrule
DEVISE~\cite{frome2013devise}  & 56.5 & 52.0 & 54.2 & 59.7 & 39.8 \\
% CONSE~\cite{norouzi2013zero} & 38.8 & 34.3 & 45.6 & 44.5 & 26.9 \\
% CMT~\cite{socher2013zero} & 39.9 & 34.6 & 39.5 & 37.9 & 28.0 \\
% IAP~\cite{lampert2013attribute} & 19.4 & 24.0 & 35.9 & 35.9 & 36.6 \\
ALE~\cite{akata2015label} & 58.1 & 54.9 & 59.9 & 62.5 & 39.7 \\
SYNC~\cite{changpinyo2016synthesized} & 56.3 & 55.6 & 54.0 & 46.6 & 23.9 \\
SAE~\cite{kodirov2017semantic} & 40.3 & 33.3 & 53.0 & 54.1 & 8.3  \\
% DEM~\cite{zhang20 17learning} & 61.9 & 51.7 & \textbf{68.4} & 67.1 & 35.0 \\
RN~\cite{sung2018learning} & - & 55.6 & 68.2 & 64.2 & - \\
% CVAE~\cite{mishra2018generative} & 61.7 & 52.1 & 71.4 & 65.8  \\
GAFE~\cite{ijcai2019-graph-auto} & 62.2 & 52.6 & 67.9 & 67.4 & \textbf{44.3} \\
% GDAN~\cite{huang2018generative} &  \\
% GMN~\cite{sariyildiz2019gradient} &  \\
% VSE~\cite{zhu2019generalized} &  \\
% PQZSL~\cite{Li2019CompressingUI} & \\
% MLSE~\cite{ding2019marginalized} & 62.8 & 64.2 & - & 67.8   \\
\midrule
APNet(ours) & \textbf{62.3} & \textbf{57.7}  & 68.0  & \textbf{68.0} &  41.3 \\
\bottomrule
\end{tabular}
\label{table:results-zs}
\end{table*}

\subsection{Evaluation Criterion}
To mitigate the bias caused by imbalance of test data for every class, following the most recent works~\cite{xian2018zero}, we evaluate the APNet's performance according to averaged per-class accuracy:
\begin{align}
 ACC_{\mathcal{Y}}=&\frac{1}{|\mathcal{Y}|}\sum_{y \in \mathcal{Y}}\frac{1}{|D_y|} \sum_{(\vx, y)\in D_y} \mathbbm{1}[\widehat{y}=y],
% \notag &D_y\triangleq\{(\vx_i,y_i)\in \mathcal D:y_i=y\}.  
\end{align}
% $ ACC_{\mathcal{Y}}=&\frac{1}{|\mathcal{Y}|}\sum_{y \in \mathcal{Y}}\frac{1}{|D_y|} \sum_{(\vx, y)\in D_y} \mathbbm{1}[\widehat{y}=y],
% \notag &D_y\triangleq\{(\vx_i,y_i)\in \mathcal D:y_i=y\}. $

% $acc_{\mathcal{Y}} =\frac{1}{\|\mathcal{Y}\|}\sum_{c \in \mathcal{Y}} \frac{\sum_{(\vx^{te}, y) \in \mathcal{D}^{c}} \mathbbm{1}[\widehat{y}=y]}{\|\mathcal{D}^{c}\|}$,
where $\mathcal{D}_{y}$ is the dataset of the data-label pairs for class $y$, and $\widehat{y}$ is the prediction of the image feature representation $\vx$.
We measured the overall accuracy of the data in the seen and unseen classes in terms of the harmonic mean of the per-class seen accuracy $ACC_{\mathcal{Y}^{seen}}$ and unseen accuracy $ACC_{\mathcal{Y}^{unseen}}$ following previous works~\cite{xian2018zero}:
$H = \frac{2*ACC_{\mathcal{Y}^{seen}}*ACC_{\mathcal{Y}^{unseen}}}{ACC_{\mathcal{Y}^{seen}}+ACC_{\mathcal{Y}^{unseen}}}$.

% where $\mathcal{D}^{c}$ is the dataset of data-label pair ($\vx$, $y$) for class $c$ and $\widehat{y}$ is the prediction for image feature representation $\vx$. 

% \begin{equation}
% acc_{\mathcal{Y}} = \frac{1}{\|\mathcal{Y}\|} \sum_{c=1}^{\|\mathcal{Y}\|} \frac{\[ \mathbbm{1} \]}{\text{\# samples in c}}
% \end{equation}

% To measure the overall accuracy of the data from the seen classes and the unseen classes, we use the harmonic mean of the per-class accuracy for both the seen classes and the unseen classes following previous settings~\cite{xian2018zero}: 

\subsection{Experimental Results}

\noindent\textbf{Generalized Zero-shot Learning.}

\noindent With generalized zero-shot learning, the setting is to classify samples ($\vx$, $y \in \mathcal{Y}^{seen}\cup\mathcal{Y}^{unseen}$) from both seen and unseen classes while training on samples ($\vx$, $y \in \mathcal{Y}^{seen}$) from seen classes. The comparison results for APNet and other baselines are shown in \Tabref{table:results-gzs}.

Classical zero-shot learning methods, e.g., CMT, CONSE, usually suffer from the imbalance problem between the training and testing stages. Their models perform well on the seen classes but the per-class accuracy on the unseen classes is low, with some models even achieving close to 0\% per-class accuracy.
It is challenging to achieve competitive results on every dataset for the generalized zero-shot learning due to the imbalanced accuracy for seen classes and unseen classes because of model overfitting to samples from the seen classes.
Our APNet outperforms state-of-the-art results on all five datasets.
Our model achieves better results especially on datasets extracted from ImageNet, i.e., CUB, AWA1, AWA2, and achieves over 3 points improvements on the overall criterion ($H$) for both seen and unseen accuracy, and
consistent improvements on the unseen accuracy of up to $\sim 3\%$.

% The more practical setting of generalized zero-shot learning is to classify samples ($\vx^{te}$, $y \in \mathcal{Y}^{seen}\cup\mathcal{Y}^{unseen}$) from both seen and unseen classes while training on samples ($\vx^{tr}$, $y \in \mathcal{Y}^{seen}$) from seen classes. The comparison results for our APN and other baselines are shown in Table.~\ref{table:results-gzs}. Classical zero-shot learning methods usually suffer from the imbalance/bias problem between the training stage and the test stage. Their model performs well on the seen classes but the per-class accuracy on the unseen classes dramatically drops and some models even perform close to 0\% per-class accuracy.
% Our APN outperforms state-of-the-art results on ?? datasets. Similar to the performance in the zero-shot learning setting, our model achieves better results on datasets extracted from ImageNet, i.e., CUB, AWA1, AWA2, and achieves over 3 improvements on the overall criterion ($H$) for both seen and unseen accuracy and consistent improvement on the unseen accuracies of up to $\sim 3\%$. 

\noindent\textbf{Zero-shot Learning.}

\noindent The setting with zero-shot learning is to classify samples ($\vx$, $y \in \mathcal{Y}^{unseen}$) from unseen classes while training on samples ($\vx$, $y \in \mathcal{Y}^{seen}$) from seen classes. The comparison results for APNet and the other baselines are shown in~\Tabref{table:results-zs}.
APNet achieves competitive results in this setting.
Similar to the results in GZSL, APNet is more powerful on subsets of ImageNet datasets, i.e., CUB, AWA1, AWA2, where APNet achieves up to $2\%$ accuracy improvements compared to the most recent baselines.

Comparing the results for both generalized zero-shot learning in Table~\ref{table:results-gzs} and zero-shot learning in Table~\ref{table:results-zs} settings, it is challenging to perform well on both settings. Methods that perform well on zero-shot learning do not guarantee a good performance on generalized zero-shot learning.
DEM~\cite{zhang2017learning} achieved similar performance compared to our APNet on AWA1. However, the generalization ability of DEM, which is measured by the performance in the setting of generalized zero-shot learning, has space for improvements. This is probably because DEM treats visual space as the embedding space for nearest neighbor search. Comparatively, we mainly do transformation over the semantic embeddings by propagation on a shared graph of training and test classes and use a parametric KNN classifier for a semantic-visual joint similarity prediction. 
The visual space used in DEM has high bias and non-i.i.d. problems and might be difficult to generalize when considering predictions over both seen classes and unseen classes.
GAFE~\cite{wang2017mgae} achieved better performance on aPY dataset but also suffers from the generalization ability problem. Similar to DEM, they also focused on the visual space and proposed a reconstruction regularizer on the visual feature representations. In remains open if the regularizer can generalize on the unseen classes without their visual feature representations during training stage.

% setting especially in subsets of ImageNet datasets, i.e., CUB, AWA1, AWA2, where APN achieves up to $2\%$ accuracy improvements compared to the most recent baselines.

% The setting of zero-shot learning is to classify samples ($\vx^{te}$, $y \in \mathcal{Y}^{unseen}$) from unseen classes while training on samples ($\vx^{tr}$, $y \in \mathcal{Y}^{seen}$) from seen classes. The comparison results for our APN and other baselines are shown in Table.~\ref{table:results-zs}. 
% APN achieves competitive results in this setting, especially in subsets of ImageNet datasets, i.e., CUB, AWA1, AWA2, where APN achieves up to $2\%$ accuracy improvements compared to the most recent baselines.
% This is because 1) we use ImageNet pretrained backbone so that the features of similar objects in ImageNet can be better extracted compared to other datasets, e.g. SUN, a scene dataset. 2) The hierachy for objects in ImageNet is clearly defined (WordNet), the clear hierarchy relationship between classes/nodes is more suitable and meaningful for attributes propagation between nodes. 

\noindent\textbf{Ablations and Variants of APNet.}

\noindent We did ablation study and developed some variants of APNet in~\Tabref{table:ablation}.
% Ablation study is on the component of propagation module.
In the ablation of propagation ($\times$ in ``Graph Hierarchy''), we skip the propagation step and directly use the node feature representations for similarity comparison and prediction.
Propagation can achieve above $1$ point improvement on $H$ under different training strategies.

Two variants of APNet are developed by adjusting the training strategy to traditional minibatch training (denoted by $\times$ in ``Meta training'') and replacing the graph hierarchy used in propagation with the hierarchy defined in WordNet.
The meta-learning style training strategy we used can bring $2 \sim 3\%$ accuracy improvements on unseen accuracy and $2 \sim 3$ points improvement on $H$ while keeping the rest components the same, which verifies that training on different tasks in every iteration can get better model generality.
When the hierarchy is defined by WordNet, the propagation graph is assumed to be fully connected and the value of the adjacency matrix for node $i$ and node $j$ ($A_{ij}$) is defined as $1/d_{ij}$, where $d_{ij}$ is the number of hops between node $i$ and node $j$ on WordNet.
Predefined graph hierarchy can bring improvements but is sensitive to the training strategy.
The reason is probably that the exact/optimal formulation for generating the adjacency matrix based on the distance between nodes is unclear even though intuitively closer nodes should have higher values for its  adjacency matrix.
Dedicated designs on how to integrate the hierarchy information more effectively have the potential to boost the performance.

\begin{table}[t!]
\centering
\caption{
Ablations and performance comparisons on variants of APNet on AWA2.
% ``S'' and ``U'' indicate per-class seen accuracy (\%) and unseen accuracy (\%), respectively. ``H'' indicates the harmonic mean of ``S'' and ``U''.
}
% \vspace{2mm}
 \resizebox{.95\columnwidth}{!}{
\begin{tabular}{cc|ccc}
\toprule
\textbf{Meta Training} & \textbf{Graph Hierarchy}  & \textbf{S}  & \textbf{U} & \textbf{H} \\
\midrule
$\times$                  &   $\times$                & 82.8        &  50.2      & 62.5       \\
$\times$                  &   WordNet                 & 83.9        &  50.4      & 62.9       \\
$\times$                  &   Learned               & 81.2        &  52.6      & 63.9       \\
$\checkmark$              &    $\times$               & \textbf{84.4}        &  53.4      & 65.4       \\
$\checkmark$              &   WordNet                 & 82.6        &  \textbf{55.7}      & \textbf{66.6}       \\
$\checkmark$              &   Learned              & 83.9        &  54.8      & \textbf{66.4}       \\
\bottomrule
\end{tabular}}
% \vspace{-2mm}
\label{table:ablation}
\end{table}

% \section{Conclusion}
% In this work, we proposed to integrate categraph structure information into the attribute representation for zero-shot learning. 
% % Structural information is integrated via an attribute propagation network (APNet), a graph propagation model that generates an attribute vector for each class, and categorizes an image to the class with the nearest attribute vector to the image's embedding.
% Structural information is integrated via an attribute propagation network (APNet).
% APNet learns to propagate attributes of every class and categorizes an image to the class with the nearest attribute vector to the image's embedding.
% Our model does not need extra hierarchy information and can generate the graph edges based on attention mechanism.
% We found that a meta-learning style training strategy can improve the generality of APNet under the setting of zero-shot learning.
% In experiments with two zero-shot learning scenarios and five benchmark datasets, APNet displayed either compelling performance or set new state-of-the-art results.

\section{ Acknowledgments}
This research was funded by the Australian Government through the Australian Research Council (ARC) under grants 1) LP160100630 partnership with Australia Government Department of Health and 2) LP180100654 partnership with KS Computer. We also acknowledge the support of NVIDIA Corporation and Google Cloud with the donation of GPUs and computation credits.

\bibliographystyle{aaai}
\bibliography{aaai}

\end{document}